\pdfoutput=1

\documentclass[11pt]{article}

\usepackage[preprint]{acl}

\usepackage{times}
\usepackage{latexsym}

\usepackage[T1]{fontenc}

\usepackage[utf8]{inputenc}

\usepackage{microtype}

\usepackage{inconsolata}

\usepackage{comment}
\usepackage{graphicx}          
\usepackage[absolute]{textpos} 
\usepackage{amsmath}
\usepackage[linesnumbered,ruled]{algorithm2e}
\usepackage{hyperref}
\usepackage{multirow}
\usepackage{booktabs}
\usepackage{subcaption}

\newcommand*{\affaddr}[1]{#1} 
\newcommand*{\affmark}[1][*]{\textsuperscript{#1}}
\newcommand*{\email}[1]{\texttt{#1}}

\title{SemEval-2025 Task 5: LLMs4Subjects - LLM-based Automated Subject Tagging for a National Technical Library's Open-Access Catalog}

\author{%
Jennifer D'Souza, Sameer Sadruddin, Holger Israel, Mathias Begoin, and Diana Slawig\\ 
\affaddr{TIB Leibniz Information Centre for Science and Technology, Hannover, Germany}\\
\email{\{firstname.lastname\}@tib.eu}%
}

\begin{document}
\maketitle

\setlength{\TPHorizModule}{\textwidth} 
\setlength{\TPVertModule}{\textwidth}  
\begin{textblock}{0.5}(0.99,0.03)
\includegraphics[scale=.3]{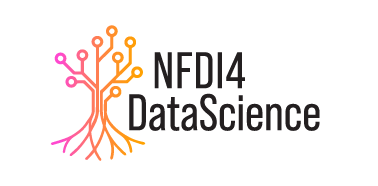}
\end{textblock}

\begin{abstract}

We present SemEval-2025 Task 5: LLMs4Subjects, a shared task on automated subject tagging for scientific and technical records in English and German using the GND taxonomy. Participants developed LLM-based systems to recommend top-k subjects, evaluated through quantitative metrics (precision, recall, F1-score) and qualitative assessments by subject specialists. Results highlight the effectiveness of LLM ensembles, synthetic data generation, and multilingual processing, offering insights into applying LLMs for digital library classification.

\end{abstract}

\section{Introduction}

Subject classification within library systems involves organizing books and resources based on their content and subject matter to facilitate easy retrieval and access. Automated methods for classifying scientific texts include Springer Nature's CSO classifier \citep{salatino2019cso}, which uses syntactic and semantic analysis to categorize papers based on the Computer Science Ontology (CSO) using abstracts, titles, and keywords. Other approaches have utilized citation metadata for classification \citep{mahdi2011automatic}, while research like New Mexico State University's application of topic modeling on digital news releases \citep{glowacka2022applying} demonstrates the diversity of techniques. Methods range from embeddings \citep{buscaldi2017classification,ruuvzivcka2021towards} and deep learning \citep{ahanger2022novel} to citation analysis \citep{small2014identifying} and topic modeling \citep{bolelli2009topic,griffiths2004finding}. Thus, while the use of NLP in digital library subject classification is well-established \citep{gooding2019subjectifying}, the potential for leveraging Large Language Models (LLMs) for their extensive knowledge representation remains largely unexplored.

Several open-source toolkits integrate machine learning (ML) and NLP for automated subject indexing, notably ANNIF (\url{https://annif.org/}), developed by the National Library of Finland, as a DIY automated subject indexing tool supporting the training of multiple traditional machine learning algorithms \cite{suominen2019annif}. ANNIF allows users to train models on a chosen subject taxonomy and metadata to generate subject headings for new documents. It has performed well on scientific papers and books but struggles with older or diverse materials like Q\&A pairs or Finnish Wikipedia. European national libraries, including Sweden's \href{https://www.kb.se/}{National Library}, the \href{https://www.zbw.eu/en/about-us/knowledge-organisation/automation-of-subject-indexing-using-methods-from-artificial-intelligence}{Leibniz Information Centre for Economics}, and the \href{https://groups.google.com/g/annif-users/c/KVQB-hvLrbA/m/I9RwM9EPBgAJ}{German National Library}, have adopted ANNIF. Supporting multiple languages and vocabularies, it offers command-line, web, and REST API interfaces, demonstrating the adaptability required for effective subject classification.

With these insights, as a SemEval 2025 shared task, we organized Task 5 — LLMs4Subjects — to explore the untapped potential of LLMs for subject classification and tagging. 
The task was defined on the catalog of the \href{https://www.tib.eu/en/tib/profile}{TIB – Leibniz Information Centre for Science and Technology}, Germany’s national library for science and technology. Its catalog, TIBKAT, holds 5.7 million records (as of March 2025), including bibliographic data and metadata from freely available electronic collections. A subset of around 100,000 records is available as  \href{https://www.tib.eu/en/services/open-data}{open access}, and the task focused on this subset. The collection includes various record types such as technical reports, publications, and books, primarily in English and German. Regardless of full-text language, records are consistently annotated using subject terms from the Gemeinsame Normdatei (GND), the integrated authority file and subject taxonomy used in the German library system.
LLMs offer promising opportunities for subject classification through their ability to process natural language at scale and capture the nuances of complex, interdisciplinary topics. This can significantly improve the accuracy and efficiency of organizing large collections, enhancing the accessibility and discoverability of information. The solutions developed in this task serve as a benchmark for applying LLMs in digital library systems, fostering innovation and setting new standards in the field. Moreover, the task aligns with the goals of the SemEval series by evaluating a novel application of computational semantics essential for effective information organization and retrieval.

While the shared task focused on practical systems development, it was also driven by the following four research questions. \textbf{RQ1} Multilingual vs. Monolingual Models in Subject Tagging: How do multilingual pre-trained models compare to monolingual models in bilingual subject tagging tasks?, \textbf{RQ2} Effect of Training Data Size and Diversity: How does the size and diversity of training data affect LLM performance in subject tagging?, \textbf{RQ3} Role of Augmented Generation: How impactful are RAG approaches versus finetuning?, and \textbf{RQ4} Efficiency of Language Models: How effective are small versus large LMs in performing the task?

A key observation from the systems submitted to this shared task is that the advantages of LLMs over traditional machine learning algorithms for subject indexing remain debatable \cite{kluge-kahler-2024-shot}. While the first iteration of this shared task brings this question to light, to further explore the possibilities offered by LLMs, we will reorganize the LLMs4Subjects shared task \href{https://sites.google.com/view/llms4subjects-germeval/}{a second time} and this time with a theme to build solutions based on \textit{energy- and compute-efficient LLMs}.

\section{Background}

\noindent{\textbf{About TIB.}} TIB – the Leibniz Information Centre for Science and Technology and University Library – promotes free access to knowledge, information sharing, and open scientific publications and data. As Germany’s national library for science and technology, including Architecture, Chemistry, Computer Science, Mathematics, and Physics, it maintains a globally unique collection, including audiovisual media and research data.


\noindent{\textbf{About the TIBKAT Open-access Subset.}} A subset of TIB’s collection—including bibliographic data in science and technology from its library catalog (TIBKAT Data), metadata from freely available electronic collections, and metadata with thumbnails from the TIB AV-Portal—is made available under the CC0 1.0 Universal Public Domain Dedication, allowing unrestricted use. More \href{https://www.tib.eu/en/services/open-data}{here}.

\noindent{\textbf{The GND Taxonomy.}} The Gemeinsame Normdatei\footnote{\url{https://www.dnb.de/EN/gnd}} (GND, German for “integrated authority file”) is an international \href{https://en.wikipedia.org/wiki/Authority_control}{authority file} used primarily by German-speaking library systems to catalog and link information on topics, organizations, people, and works. It is publicly available for download\footnote{The GND is available for download at \url{https://www.dnb.de/EN/Professionell/Metadatendienste/Datenbezug/Gesamtabzuege/gesamtabzuege_node.html}.} in various formats under a CC0 license. The GND's records specifically cover entities such as persons, corporate bodies, conferences, geographic locations, subject headings, and works, relevant to cultural and scientific collections. 

For the LLMs4Subjects shared task, only the GND subject heading (Sachbegriff) records are of interest. Since accessing the GND for the first time for new users can be overwhelming, for the convenience of our participants, we have created a \href{https://github.com/jd-coderepos/llms4subjects/tree/main/supplementary-datasets/gnd-how-to}{how-to guide} to download the latest GND file.

\section{Source Dataset}

We queried the TIBKAT service to restrict its metadata to records containing abstracts and GND subject indexing. The query is fully reproducible via this \href{https://www.tib.eu/en/search?tx_tibsearch_search%5Baction%5D=search&tx_tibsearch_search%5Bcnt%5D=20&tx_tibsearch_search%5Bcontroller%5D=Search&tx_tibsearch_search%5BgroupField%5D=matchTitleTypeFirstAuthor_str&tx_tibsearch_search%5Bquery%5D=prefix%3Atibkat%20%2Babstract%3A%2A%20%2BxmlPath%3Asubject%2F%40type%3Dgnd&cHash=90eaa41371219f0a5764591225498b6d}{persistent search link}. It returned 189,665 records at the time of dataset creation. The TIB open-access catalog spans nine media types: Book (136,434), Thesis (31,859), Conference (12,212), Report (6,711), Article (2,080), Collection (188), AudioVisualDocument (167), Periodical (57), and Chapter (11), detailed in \autoref{sec:mt-def}. Using the \texttt{langdetect}\footnote{\url{https://pypi.org/project/langdetect/}} Python library, we identified 48 languages. The top five were German (108,637), English (76,735), French (1,741), Indonesian (945), and Spanish (311).\footnote{Reflecting the real-world nature of the corpus, many records contain mixed-language content and are not reliably classifiable under a single language.} For the official shared task corpus, we retained only records in German and English and excluded the four least represented media types, resulting in a dataset of 123,589 records. The excluded data is available as \href{https://github.com/jd-coderepos/llms4subjects/tree/main/supplementary-datasets/tibkat-supplementary-dataset}{supplementary data}. The final shared task dataset is available at: \url{https://github.com/jd-coderepos/llms4subjects/tree/main/shared-task-datasets}.

\begin{table*}[!htb]
\centering
\begin{tabular}{|p{2.5cm}|l|l|l|l|l|l|} \hline
statistics & lang & Article & Book   & Conference & Report & Thesis \\ \hline
\multirow{2}{*}{num. records}  & en   & 1,042/253 & 26,966/17,669 & 3,619/2,840  & 1,275/896  & 3,452/2,506  \\ \cline{2-7}
 & de   & 6/5 & 33,401/12,528 & 2,210/717 & 1,507/761 & 8,459/3,727 \\ \hline
\multirow{2}{2.5cm}{num. subjects (avg, max)}   & en   &  (3/4, 7/6) & (3/3, 39/26) & (3/3, 14/16) & (3/3, 12/13) & (4/4, 20/19) \\ \cline{2-7}
    & de   & (3/3, 8/7) & (3/3, 27/25) & (3/4, 17/16) & (3/3, 15/15) & (4/4, 20/19) \\ \hline
\end{tabular}
\caption{Train dataset statistics (\href{https://github.com/jd-coderepos/llms4subjects/tree/main/shared-task-datasets/TIBKAT/all-subjects}{all-subjects}/\href{https://github.com/jd-coderepos/llms4subjects/tree/main/shared-task-datasets/TIBKAT/tib-core-subjects}{tib-core} collections) for the LLMs4Subjects shared task.}
\vspace{-10pt}
\label{tab:dataset-stats}
\end{table*}

\subsection{How are subject annotations obtained?}

Subject annotations in the TIB catalog are continuously created by a dedicated team of 17 expert subject specialists covering 28 disciplines—including Architecture, Chemistry, Electrical Engineering, Mathematics, Traffic Engineering, and \href{https://terminology.tib.eu/ts/ontologies/linsearch/individuals?iri=https%3A%2F%2Fpurl.org%2Flinsearch}{others}—ensuring broad and expert-driven subject classification.

In libraries, content is typically described using controlled vocabularies. In Germany, the GND is used for cataloging literature. In addition to descriptive cataloging (e.g., author, title, year, publisher), subject cataloging is performed by subject librarians. Based on the title, abstract, and full text, librarians assign appropriate GND keywords to describe the content as precisely as possible. This collaborative work is carried out across various libraries and national library networks.

With TIB adding around 15,000 new titles each month, subject cataloging is a labor-intensive task. Integrating AI-driven solutions—especially LLMs—can significantly boost efficiency, partially automate workflows, and improve usability, all while maintaining cataloging quality. Such innovations are key to modernizing information management and supporting research at scale.

\section{Shared Task Description and Dataset}

The LLMs4Subjects shared task challenged participants to develop LLM-based systems for recommending relevant subjects from the GND taxonomy to annotate TIB technical records. Given a record’s title and abstract as input, systems were expected to generate a customizable top-$k$ ranked list of relevant GND subjects. Since the dataset included records in both English and German, systems were required to support bilingual semantic processing. The task was defined over two dataset collections.

\subsection{Dataset Collections}

\noindent\textbf{all-subjects.}\footnote{\url{https://github.com/jd-coderepos/llms4subjects/tree/main/shared-task-datasets/TIBKAT/all-subjects}}
This dataset comprises the full TIBKAT open-source collection, with predefined splits: 81,937 records for training and 13,666 for development. A detailed dataset overview is available in the \href{https://github.com/jd-coderepos/llms4subjects/tree/main/shared-task-datasets/TIBKAT/all-subjects}{shared task repository}. Participants also received the accompanying GND subject taxonomy,\footnote{\url{https://github.com/jd-coderepos/llms4subjects/blob/main/shared-task-datasets/GND/dataset/GND-Subjects-all.json}} which included 204,739 subjects, with coverage and distribution frequencies published \href{https://github.com/jd-coderepos/llms4subjects/blob/main/shared-task-datasets/TIBKAT/all-subjects/data-statistics/subject_frequencies.csv}{online}.

Due to the large dataset size (\textgreater100,000 records), participants could opt for a smaller subset focused on TIB’s core subject classification.  

\noindent\textbf{tib-core.}\footnote{\url{https://github.com/jd-coderepos/llms4subjects/tree/main/shared-task-datasets/TIBKAT/tib-core-subjects}}
This subset includes only records annotated with at least one GND subject from the so-called TIB core domains. It contains 41,902 training and 6,980 development records across 14 domains: Architecture (arc), Civil Engineering (bau), Mining (ber), Chemistry (che), Chemical Engineering (cet), Electrical Engineering (elt), Materials Science (fer), Information Technology (inf), Mathematics (mat), Mechanical Engineering (mas), Medical Technology (med), Physics (phy), Engineering (tec), and Traffic Engineering (ver). A refined GND subject taxonomy\footnote{\url{https://github.com/jd-coderepos/llms4subjects/blob/main/shared-task-datasets/GND/dataset/GND-Subjects-tib-core.json}} with 79,427 subjects accompanied this dataset, along with \href{https://github.com/jd-coderepos/llms4subjects/blob/main/shared-task-datasets/TIBKAT/tib-core-subjects/data-statistics/subject_frequencies.csv}{subject coverage and frequency distributions}.

Participants could choose between the all-subjects dataset for comprehensive indexing, the tib-core dataset for a more focused classification task, or even attempt both.

\subsection{Dataset Format and Statistics}

Both datasets were released in JSON-LD format and include metadata such as title, type, abstract, and authors. The key attribute for the task, \href{https://www.dublincore.org/specifications/dublin-core/dcmi-terms/elements11/subject/}{\texttt{dcterms:subject}}, holds the GND subject headings assigned to each record. Example records are available in English \href{https://github.com/jd-coderepos/llms4subjects/blob/main/shared-task-datasets/TIBKAT/all-subjects/data/train/Article/en/3A1499846525.jsonld}{here} and in German \href{https://github.com/jd-coderepos/llms4subjects/blob/main/shared-task-datasets/TIBKAT/all-subjects/data/train/Article/de/3A168396733X.jsonld}{here}. \autoref{tab:dataset-stats} provides a detailed dataset breakdown. Books were the most common record type in both language collections, with each record annotated with an average of 3 to 7 subjects.

\section{Task Setup}

The shared task offered multiple communication channels: a dedicated website,\footnote{\url{https://sites.google.com/view/llms4subjects/}} a \href{https://groups.google.com/u/6/g/llms4subjects}{Google Group} for FAQs, and direct email support (llms4subjects@gmail.com). The organizing team included two Computer Scientists and three TIB subject specialists who supported participants throughout. The \href{https://sites.google.com/view/llms4subjects/important-dates?authuser=0}{task timeline} spanned four months, from October 2024 to January 2025. A \href{https://forms.gle/zRpD46iDpKJmWU3bA}{Declaration of Interest for Participation (DIP) survey} initially recorded \href{https://docs.google.com/spreadsheets/d/1RNrDGy7D_lMVm-v54nnuFJezQe78MEE4EUdU02MT5BQ/edit?usp=sharing}{\textbf{33 teams}}; of these, \textbf{14 teams} submitted system outputs, and \textbf{12} also contributed system description papers to the SemEval workshop. The next section summarizes their approaches.

\section{Shared Task Participant Systems}

The participant systems are described with a focus on their key methodological contributions. Teams are listed in alphabetical order of their names. An overview of the systems is provided in \autoref{tab:team_results}.

\begin{table*}[!htb]
    \centering
    \small
    \begin{tabular}{l p{5cm} p{4.5cm} p{3cm}}
        \toprule
        \textbf{Team} & \textbf{Method} & \textbf{LLMs Used} & \textbf{Ranking} \\
        \midrule
        Annif & LLM-based synthetic data generation and XMTC traditional classifier models ensemble & Llama-3.1-8B-Instruct & 1st all-subjects, 2nd tib-core, 4th qualitative \\
        DNB-AI & Few-shot prompting to an LLM ensemble & Llama-3.2-3B-Instruct, Llama-3.1-70B-Instruct, Mistral-7B-v0.1, Mixtral-8x7B-Instruct-v0.1, OpenHermes-2.5-Mistral-7B, Teuken-7B-instruct-research-v0.4, LLama-3.1-8B-Instruct & 4th all-subjects, N/A tib-core, 1st qualitative \\
        DUTIR831 & Synthetic data generation, GND knowledge distillation, and supervised finetuning & Qwen2.5-72B-Instruct & 2nd all-subjects, 4th tib-core, 2nd qualitative \\
        Homa & RAG-based ranking and retriever finetuning & Qwen2.5-0.5B & N/A all-subjects, 10th tib-core, 10th qualitative \\
        Jim & BERT model ensemble & German BERT, Multilingual cased/uncased, ModernBERT base & 7th all-subjects, N/A tib-core, 5th qualitative \\
        LA²I²F & Transfer of concepts from similar documents to target and concept similarity to target document & Llama-3.1-8B as baseline, all-mpnet-base-v2 Sentence Transformer & 6th all-subjects, 3rd tib-core, 8th qualitative \\
        last\_minute & Subjects are ranked, then re-ranked with embeddings, and refined with an LLM & stella-en-400M-v5, granite-embedding-125m-english, Llama-3.2-1B & N/A all-subjects, 9th tib-core, 11th qualitative \\
        NBF & Finetuned embeddings using Burst Attention and multi-layer perceptron & all-mpnet-base-v2, german-roberta & 9th all-subjects, N/A tib-core, 9th qualitative \\
        RUC Team & Retrieves pre-indexed similar records and uses their subject tags as candidates & Arctic-Embed 2.0, Llama-8B, Chat GLM 4 (130B) & 3rd all-subjects, 1st tib-core, 3rd qualitative \\
        silp\_nlp & Multilingual sentence transformer-based embedding similarity & jina-embeddings-v3-559M, distiluse-base-multilingual-cased-v2 & 11th all-subjects, 6th tib-core, N/A qualitative \\
        TartuNLP & Bi-encoder candidate subject retrieval, and finetuned cross-encoder re-ranking model & multilingual-e5-large-instruct, mdeberta-v3-base & 8th all-subjects, 7th tib-core, 7th qualitative \\
        YNU-HPCC & Combines Sentence-BERT with contrastive learning & distilroberta, minilm, mpnet & 10th all-subjects, 8th tib-core, 12th qualitative \\
        \bottomrule
    \end{tabular}
    \caption{Overview of teams, methods, LLMs used, and rankings from quantitative scores over the two dataset collections and from the qualitative evaluations. The \href{https://sites.google.com/view/llms4subjects/team-results-leaderboard}{full leaderboard} is released on our shared task website.}
    \label{tab:team_results}
\end{table*}

\noindent{\textbf{1. \href{https://github.com/NatLibFi/Annif-LLMs4Subjects/}{Annif}}} \cite{annif} The key ideas in this system's subject tagging approach were: 1) \textbf{Traditional extreme multi-label text classification (XMTC)} implemented in the Annif toolkit \citeyearpar{suominen2022annif} – Using Omikuji Bonsai \cite{khandagale2020bonsai}, a tree-based machine learning approach; MLLM (Maui-like \cite{medelyan2009human} Lexical Matching), a lexical matching algorithm; and XTransformer, a transformer-based classification model \cite{yu2022pecos} for XMTC. 2) \textbf{Ensemble Models} – Combining individual classifiers into simple averaging and neural ensembles to improve predictions. 3) \textbf{LLM-Assisted Translation} – Using the \href{https://huggingface.co/meta-llama/Llama-3.1-8B-Instruct}{Llama-3.1-8B-Instruct LLM} \citeyearpar{grattafiori2024llama} to translate bibliographic records and subject vocabularies into English and German. 4) \textbf{Synthetic Data Generation} – Expanding training data with the same LLM by generating new records with modified subject labels. And 5) \textbf{Multilingual Merging} – Combining monolingual predictions to form a multilingual ensemble, improving overall performance.

\noindent{\textbf{2. \href{https://github.com/deutsche-nationalbibliothek/semeval25_llmensemble}{DNB-AI-Project}}} \cite{dnb-ai-project} This was an LLM-driven ensemble approach which achieved top qualitative scores without fine-tuning and few-shot prompting. Key steps included: 1) \textbf{LLM Ensemble for Keyword Generation} – Multiple off-the-shelf LLMs use few-shot prompting with 8-12 examples to improve recall and precision. 2) \textbf{Map} – A BGE-M3 embedding model \cite{chen2024m3} maps LLM-generated free keywords to controlled GND subject terms via nearest neighbor search. 3) \textbf{Summarize} – Predictions from the LLM ensemble are aggregated, with similarity scores summed and normalized into a confidence-based score. 4) \textbf{Rank} – A new LLM, \href{https://huggingface.co/meta-llama/Llama-3.1-8B-Instruct}{Llama-3.1-8B-Instruct} \citeyearpar{grattafiori2024llama} assesses relevance of each predicted term on a 0-10 scale, refining rankings beyond frequency-based measures. And 5) \textbf{Combine} – Ensemble and relevance scores are weighted and combined to optimize subject ranking.

\noindent{\textbf{3. DUTIR831}} \cite{dutir831} The key steps of this system are: 1) \textbf{Data Synthesis and Filtering} – The \href{https://huggingface.co/Qwen/Qwen2.5-72B-Instruct}{Qwen2.5-72B-Instruct} LLM is used to generate synthetic data to expand training sets by selecting related subject terms and creating titles and abstracts. The LLM is then applied to filter low-quality samples based on coherence and relevance. 2) \textbf{GND Knowledge Distillation} – The LLM is then finetuned on GND subject collections improves its understanding of subject hierarchies and relationships. 3) \textbf{Supervised Fine-Tuning and Preference Optimization} – LoRA-based \cite{lora} fine-tuning on TIBKAT data is combined with Direct Preference Optimization (DPO) \cite{rafailov2023direct} to align model outputs with human-like subject assignments. 4) \textbf{Subject Term Generation} – A multi-sampling ranking strategy improves diversity, LLM-based keyword extraction selects high-confidence terms, and BGE-M3 \citeyearpar{chen2024m3} embedding-based vector retrieval adds missing terms to ensure 50 subject labels per record. And 5) \textbf{Re-Ranking for Final Selection} – Subject terms from multiple sources are re-ranked using LLMs to prioritize the most relevant terms, improving recall and ranking consistency.

\noindent{\textbf{4. Homa}} \cite{homa} Subject tagging is tackled using retrieval-augmented generation (RAG) \cite{lewis2020retrieval} to match TIBKAT records with GND subjects leveraging the OntoAligner toolkit \cite{ontoaligner}. The key methods steps are: 1) \textbf{Multi-Level Data Representation} – Records are represented at three levels: title-based, contextual (including metadata), and hierarchical (parent-level relationships) to improve subject mapping. 2) \textbf{Retrieval with Embeddings} – Nomic-AI embeddings \cite{nussbaum2024nomic} are used to retrieve the top-k relevant subjects by computing cosine similarity between records and subject embeddings. 3) \textbf{LLM-Assisted Subject Selection} – \href{https://huggingface.co/Qwen/Qwen2.5-0.5B-Instruct}{Qwen2.5-0.5B-Instruct} LLM \cite{yang2024qwen2} assesses retrieved subjects, verifying relevance in a RAG-based ranking framework. And 4) \textbf{Fine-Tuning with Contrastive Learning} – The retriever is fine-tuned using contrastive learning, training on positive and negative record-subject pairs to improve distinction between relevant and irrelevant subjects.

\noindent{\textbf{5. \href{https://github.com/jimfhahn/SemEval-2025-Task5}{Jim}}} \cite{jim} The key method steps of this system are: 1) \textbf{Multilingual BERT Ensemble} – The system uses an ensemble of four BERT models (two multilingual \href{https://huggingface.co/jimfhahn/bert-multilingual-cased}{m1} \& \href{https://huggingface.co/jimfhahn/bert-multilingual-uncased}{m2}, one \href{https://huggingface.co/jimfhahn/bert-german-cased}{German-only}, one \href{https://huggingface.co/jimfhahn/ModernBERT-base-gnd}{English-only}). 2) \textbf{Fine-Tuning on TIBKAT and GND Data} – The models are finetuned on TIBKAT records paired with GND subject labels, leveraging the AutoTrain framework \cite{thakur2024autotrain} for efficient optimization. 3) \textbf{Ensemble-Based Inference} – Subject predictions are ranked by summing confidence scores across models.

\noindent{\textbf{6. LA$^2$I$^2$F}} \cite{la2i2f} The system retrieves subjects based on document similarity (analogical reasoning) and semantic similarity with ontology concepts (ontological reasoning), combining both for optimal subject tagging. The key steps are: 1) \textbf{Embedding-Based Retrieval} – \href{https://huggingface.co/sentence-transformers/all-mpnet-base-v2}{MPNet sentence embeddings} \cite{reimers2019sentence} are used to represent documents and GND subjects in a shared vector space, enabling similarity-based matching. 2) \textbf{Analogical Reasoning for Subject Transfer} – The system identifies semantically similar training documents and transfers their human-assigned subject labels to the target document. 3) \textbf{Ontology-Based Subject Matching} – GND subjects are embedded and matched to documents based on semantic closeness, retrieving the most conceptually relevant subjects. And 4) \textbf{Final Fusion and Re-Ranking} – Predictions from both methods are merged and ranked by similarity, ensuring complementary information is integrated.

\noindent{\textbf{7. last\_minute}} \cite{last-minute} The system followed a rank, rerank, and refine approach using contextual vector embeddings stored in the Milvus vector database for efficient retrieval. The key steps are: 1) \textbf{Finetuned Embedding Model} – The stella-en-400M-v5 model \cite{zhang2024jasper} is finetuned on the training data with Multiple Negatives Ranking Loss \cite{henderson2017efficient}. 2) \textbf{Re-Ranking with a Cross-Encoder Model} – A granite-embedding-125m model \cite{granite2024embedding} re-ranks the top 100 retrieved subject tags from the prior step, refining predictions before LLM processing. And 3) \textbf{LLM Refinement} – The Llama-3.2-1B LLM \citeyearpar{grattafiori2024llama} evaluates and selects the top 50 most relevant subject tags from the re-ranked list using prompt-based filtering.

\noindent{\textbf{8. NBF}} \cite{nbf} The system introduces the use of Burst Attention \cite{sun2024burstattention}, a lightweight self-attention mechanism that treats each embedding dimension as a token, capturing inter-dimensional dependencies to enhance subject retrieval. The methodology consists of four key steps. 1) \textbf{Sentence Transformer Embeddings} \citeyearpar{reimers2019sentence}: Articles and GND subjects are embedded using \href{https://huggingface.co/sentence-transformers/all-mpnet-base-v2}{all-mpnet-base-v2} for en and \href{https://huggingface.co/T-Systems-onsite/german-roberta-sentence-transformer-v2}{german-roberta-sentence-transformer-v2} for de, aligning them in a shared space. 2) \textbf{Margin-Based Retrieval with BurstAttention}: The model is trained with a margin-based ranking loss, leveraging BurstAttention to refine embeddings by bringing relevant subjects closer and pushing irrelevant ones away. 3) \textbf{Feed-Forward MLP for Refinement}: A multi-layer perceptron (MLP) further refines embeddings to improve subject retrieval accuracy. 4) \textbf{Top-k Search}: At inference, cosine similarity between article and subject embeddings is computed, selecting the top-$k$ closest subjects as final predictions.


\noindent{\textbf{9. RUC Team}} \cite{ruc-team} This team used a retrieval-based method, prioritizing accuracy, speed, and scalability over heavy LLM inference. Key steps are: 1) \textbf{Cross-Lingual Embeddings for Retrieval} – Uses \href{https://ollama.com/library/snowflake-arctic-embed2}{Arctic-Embed 2.0} \cite{yu2024arctic}, a multilingual embedding model, to match documents across English and German in a shared semantic space. 2) \textbf{Vector-Based Nearest Neighbor Search} – Computes inner product similarities between document embeddings using Faiss indexing to efficiently retrieve the most relevant records. And 3) \textbf{Ranking Relevant Subject Terms} – Merges and re-ranks candidate subjects based on document similarity, term position, and term occurrence in the title/abstract.

\noindent{\textbf{10. silp\_nlp}} \cite{silp-nlp} This system utilizes \textbf{sentence transformer embeddings} \citeyearpar{reimers2019sentence} for titles and abstracts, retrieving subjects based on cosine similarity. It employs \href{https://huggingface.co/jinaai/jina-embeddings-v3}{JinaAi/jina-embeddings-v3} \cite{sturua2024jina}, a \textbf{novel multilingual model supporting 89 languages}, to process both en and de text. Performance was compared against \href{https://huggingface.co/sentence-transformers/distiluse-base-multilingual-cased-v2}{distiluse} sentence transformers, with JinaAi embeddings achieving superior results.

\noindent{\textbf{11. TartuNLP}} \cite{tartunlp} The system first retrieves a coarse set of candidate subjects using a \textbf{bi-encoder}, viz. \href{https://huggingface.co/intfloat/multilingual-e5-large-instruct}{multilingual-e5-large-instruct} \cite{wang2024multilingual}, then refines the selection with a \textbf{cross-encoder re-ranking model}, viz. \href{https://huggingface.co/microsoft/mdeberta-v3-base}{mdeberta-v3-base model} \cite{hedebertav3}, finetuned on the task dataset. The key insight being the two-stage approach nearly doubles recall compared to using the bi-encoder alone, confirming the cross-encoder re-ranking’s impact on performance.

\noindent{\textbf{12. \href{https://github.com/Jiangnaio/SemEval2025Task5}{YNU-HPCC}}} \cite{ynu-hpcc} The system fine-tuned multilingual sentence-BERT models, such as paraphrase-multilingual-\href{https://huggingface.co/sentence-transformers/paraphrase-multilingual-MiniLM-L12-v2}{MiniLM}, -\href{https://huggingface.co/sentence-transformers/paraphrase-multilingual-mpnet-base-v2}{mpnet}, and \href{https://huggingface.co/sentence-transformers/all-mpnet-base-v2}{mpnet-base}, using contrastive loss, to improve semantic alignment between the records and subjects. Key was their \textbf{Balanced Positive-Negative Sampling} – Two strategies were tested: multi-label sampling, aggregating all true labels per document, and single-label sampling, constructing 1:1 positive-negative pairs to improve classification.

The shared task attracted a diverse range of systems. A review of the 12 submissions revealed unique methodological contributions, including Burst Attention \cite{nbf}, analogical/ontological reasoning \cite{la2i2f}, the use of toolkits like Annif \citeyearpar{suominen2022annif} \cite{annif} and OntoAligner \citeyearpar{ontoaligner} \cite{homa}, and multi-stage prompt engineering. Some teams \cite{ruc-team,silp-nlp} also evaluated newer embedding models such as Arctic-Embed and JinAI. As shown in \autoref{tab:team_results}, top-performing systems went beyond standard sentence transformer embeddings \citeyearpar{reimers2019sentence}. Key strategies among the leading teams—Annif \cite{annif}, DNB-AI \cite{dnb-ai-project}, DUTIR831 \cite{dutir831}, RUC Team \cite{ruc-team}, and Jim \cite{jim}—included: (1) model ensembles, (2) synthetic training data generation, and (3) multilingual language models, the latter being common across submissions. Notably, DUTIR831 and RUC Team deployed very large LLMs—Qwen2.5-72B-Instruct and ChatGLM 4 (130B)—while others used models with 8B or fewer parameters.

In the next section, we present the leaderboard results for the shared task.

\begin{table*}[!tb]
    \centering
    \small
    \begin{tabular}{p{1.6cm} c c c c c | p{1.6cm} c c c c c}
        \toprule
        \multicolumn{6}{c|}{\textbf{all-subjects}} & \multicolumn{6}{c}{\textbf{tib-core}} \\
        \midrule
        Team Name & P@5 & R@5 & P@10 & R@10 & Ov. R@k & Team Name & P@5 & R@5 & P@10 & R@10 & Ov. R@k \\
        \midrule
        Annif            & 0.26 & 0.49 & 0.16 & 0.57 & 0.63 & RUC Team        & 0.25 & 0.48 & 0.16 & 0.57 & 0.66 \\
        DUTIR831         & 0.26 & 0.48 & 0.15 & 0.56 & 0.6 & Annif           & 0.23 & 0.48 & 0.14 & 0.54 & 0.59 \\
        RUC Team        & 0.23 & 0.44 & 0.14 & 0.52 & 0.59 & LA$^2$I$^2$F          & 0.2 & 0.41 & 0.13 & 0.49 & 0.58 \\
        DNB-AI  & 0.25 & 0.47 & 0.15 & 0.54 & 0.56 & DUTIR831        & 0.23 & 0.49 & 0.13 & 0.54 & 0.56 \\
        icip            & 0.2 & 0.39 & 0.12 & 0.46 & 0.53 & icip            & 0.17 & 0.37 & 0.1 & 0.44 & 0.5 \\
        LA$^2$I$^2$F          & 0.17 & 0.34 & 0.11 & 0.41 & 0.48 & silp\_nlp & 0.16 & 0.34 & 0.11 & 0.42 & 0.49 \\
        Jim             & 0.18 & 0.34 & 0.11 & 0.41 & 0.47 & TartuNLP        & 0.14 & 0.3 & 0.09 & 0.36 & 0.4 \\
        TartuNLP        & 0.13 & 0.27 & 0.08 & 0.33 & 0.38 & YNU-HPCC              & 0.05 & 0.12 & 0.03 & 0.16 & 0.23 \\
        NBF             & 0.08 & 0.17 & 0.06 & 0.23 & 0.32 & last\_minute    & 0.02 & 0.05 & 0.02 & 0.08 & 0.21 \\
        YNU-HPCC              & 0.04 & 0.09 & 0.03 & 0.12 & 0.17 & Homa            & 0.08 & 0.15 & 0.05 & 0.18 & 0.2 \\
        silp\_nlp & 0.05 & 0.08 & 0.03 & 0.11 & 0.13 & TSOTSALAB       & 0.02 & 0.04 & 0.01 & 0.05 & 0.07 \\
        Homa            & -      & -      & -      & -      & -      & NBF            & -      & -      & -      & -      & -      \\
        TSOTSALAB       & -      & -      & -      & -      & -      & DNB-AI & -      & -      & -      & -      & -      \\
        last\_minute    & -      & -      & -      & -      & -      & Jim            & -      & -      & -      & -      & -      \\
        \bottomrule
    \end{tabular}
    \caption{Quantitative performance comparisons for all-subjects and tib-core datasets. The `Ov. R@k' column represents the average recall across $k = 5, 10, ..., 45, 50$.}
    \label{tab:overall-quant-results}
\end{table*}

\begin{figure*}[!tb]
    \centering
    \small
    \begin{subfigure}{\textwidth}
        \centering
        \includegraphics[width=\linewidth]{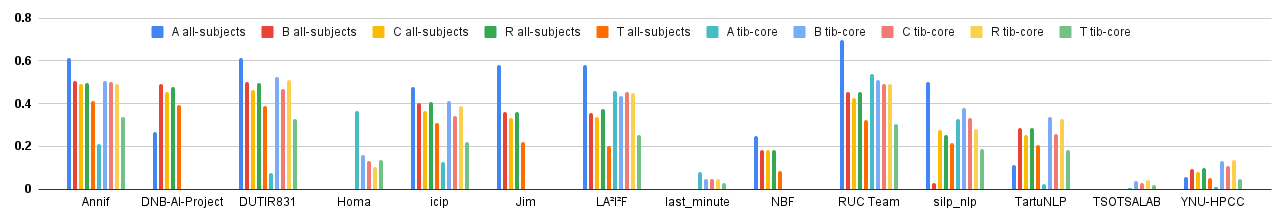}
        \caption{Recall@5 results for all-subjects and tib-core datasets based on the record-type ablation.}
        \label{fig:rec-recordtype}
    \end{subfigure}


    \begin{subfigure}{\textwidth}
        \centering
        \includegraphics[width=\linewidth]{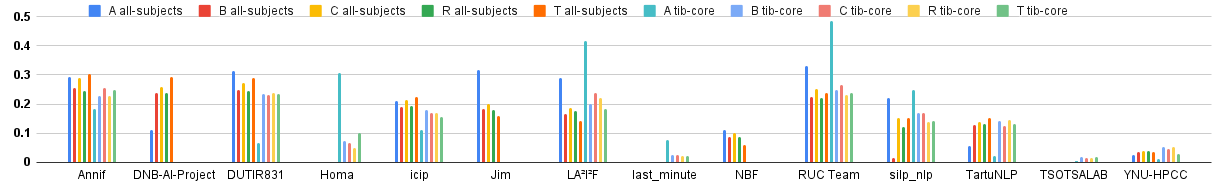}
        \caption{Precision@5 results for all-subjects and tib-core datasets based on the record-type ablation.}
        \label{fig:prec-recordtype}
    \end{subfigure}

    \caption{K@5 Results on the record type ablation. A - article, B - book, C - conference, R - report, and T - thesis. On the x-axis, teams are listed in alphabetical order of names.}
    \label{fig:recordtype-quant-results}
\end{figure*}

\begin{figure*}[!htb]
    \centering
    \small
    \begin{subfigure}{\textwidth}
        \centering
        \includegraphics[width=\linewidth]{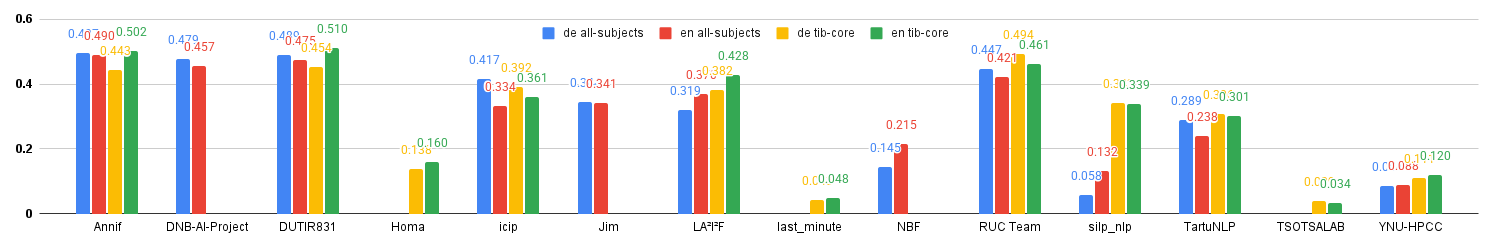}
        \caption{Recall@5 results for all-subjects and tib-core datasets based on the language ablation.}
        \label{fig:rec-lang}
    \end{subfigure}


    \begin{subfigure}{\textwidth}
        \centering
        \includegraphics[width=\linewidth]{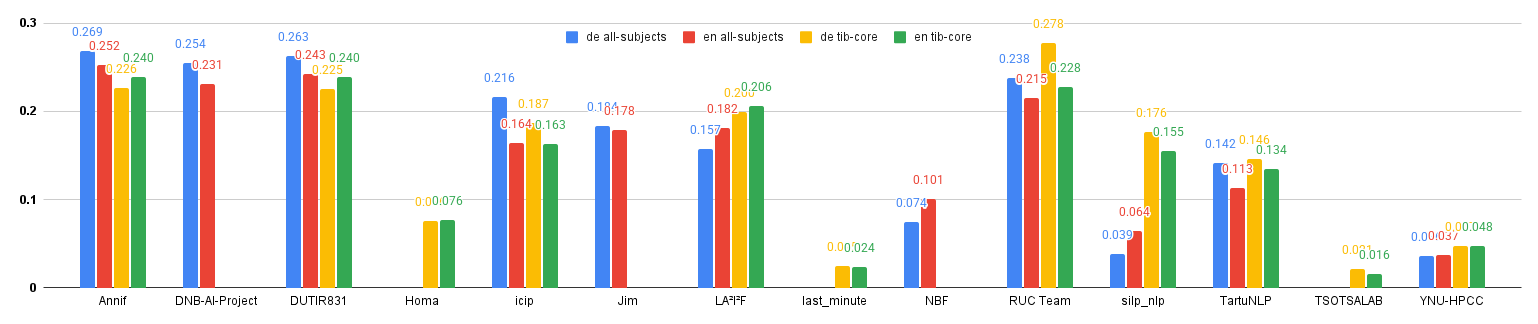}
        \caption{Precision@5 results for all-subjects and tib-core datasets based on the language ablation.}
        \label{fig:prec-lang}
    \end{subfigure}

    \caption{K@5 Results on the language ablation. On the x-axis, teams are listed in alphabetical order of names.}
    \label{fig:lan-quant-results}
\end{figure*}

\section{Shared Task Leaderboard Results}

In this shared task, we provided two leaderboards: 1) quantitative results and 2) qualitative results.

\noindent{\textbf{Quantitative Metrics.}} System performance was evaluated using average precision@k, recall@k, and F1-score@k at multiple cutoffs (k = 5, 10, 15, ..., 50). These metrics were chosen as subject tagging was treated as a bag-of-words among applicable subjects, making precision, recall, and F1-score more suitable. Given the dataset structure of the LLMs4Subjects shared task, evaluation scores were released at varying levels of granularity: (1) language-level, separately for en and de, (2) record-level, across five types of technical records, and (3) combined language and record-levels, offering a comprehensive performance breakdown. This approach provided deeper insights into system performance and facilitated detailed discussions in the task overview and system description papers. To ensure transparency, the shared task evaluation script was publicly released.\footnote{\url{https://github.com/jd-coderepos/llms4subjects/blob/main/shared-task-eval-script/llms4subjects-evaluation.py}}

\noindent{\textbf{Qualitative Metrics.}} To assess system-generated results in real-world scenarios, a qualitative evaluation was conducted over three weeks. TIB subject specialists manually reviewed 122 test records common to both all-subjects and tib-core, sampling 10 records from each of 14 subject classifications. The top 20 GND codes from teams' submissions were extracted, and subject librarians labeled them as Y (correct), I (irrelevant but technically correct), or N/Blank (incorrect). Two evaluation criteria were used: case 1 - treating both Y and I as correct, and case 2 - considering only Y. Results were summarized using average P@20, R@20, and F1@20.

Detailed results leaderboards are released on the shared task website.\footnote{\url{https://sites.google.com/view/llms4subjects/team-results-leaderboard}}

\subsection{Quantitative Evaluations}

The primary evaluation metric was recall, with the overall leaderboard ranking based on average recall scores across k values from 5 to 50. For practical use by subject specialists, systems should predict relevant subjects at lower k values, ideally between 5 and 10, with a maximum of 20. \autoref{tab:overall-quant-results} presents the results for both collections: all-subjects and tib-core. The top teams consistently predicted over half of the subject annotations in both collections. A caveat here is that our precision score would never amount to one and heavily penalizes actual system performance.\footnote{Precision@k was computed as the number of correct predictions among the top-$k$, divided by $k$. Since TIBKAT records contain on average around 5 true GND subjects, this limits how high precision can go at larger $k$. For example, even if a system perfectly predicts all 5 true subjects, the maximum possible precision at $k=10$ would still be $0.5$, since at most 5 out of 10 predictions can be correct. Therefore, for higher values of $k$, precision will necessarily be less than 1.} Despite this, they were included to provide a comparison. The top three teams based on average recall for all-subjects were Annif, DUTIR831, and RUC Team, while for tib-core, the top three were RUC Team, Annif, and LA$^2$I$^2$F. Notably, the top-performing teams on all-subjects maintained strong rankings on tib-core, with DUTIR831 placing fourth. LA$^2$I$^2$F, ranking third on tib-core, appeared more effective on the smaller subjects taxonomy of tib-core.  

At the record-type level (\autoref{fig:rec-recordtype}), most systems achieved high recall@5 for articles in both collections, while books, conference papers, and reports showed similar performance. The weakest results were observed for theses. For the top teams on all-subjects (Annif, DUTIR831, and RUC Team), precision scores across record types were similar (\autoref{fig:prec-recordtype}). RUC Team demonstrated consistently high precision on articles in both collections. On tib-core, LA$^2$I$^2$F’s boost to third place stemmed from its high precision on articles—second only to RUC—despite comparable recall scores to other teams. At the language level, results from both recall (\autoref{fig:rec-lang}) and precision (\autoref{fig:prec-lang}) showed no significant difference between processing de or en records across all teams. The only consistent variation was that RUC Team performed slightly better on de records, while LA$^2$I$^2$F for en records.

\begin{table*}[!tb]
    \centering
    \small
    \begin{tabular}{p{1.8cm} c c c c c | p{1.8cm} c c c c c}
        \toprule
        \multicolumn{6}{c|}{\textbf{qualitative eval. case 1}} & \multicolumn{6}{c}{\textbf{qualitative eval. case 2}} \\
        \midrule
        Team Name & P@5 & R@5 & P@10 & R@10 & Ov. R@k & Team Name & P@5 & R@5 & P@10 & R@10 & Ov. R@k \\
        \midrule
        DNB-AI & 0.74 & 0.33 & 0.65 & 0.54 & 0.57 & DNB-AI & 0.53 & 0.34 & 0.41 & 0.5 & 0.51 \\
        DUTIR831       & 0.7  & 0.31 & 0.61 & 0.49 & 0.53 & DUTIR831       & 0.49 & 0.32 & 0.39 & 0.46 & 0.49 \\
        RUC Team       & 0.71 & 0.28 & 0.6  & 0.46 & 0.52 & RUC Team       & 0.48 & 0.29 & 0.38 & 0.43 & 0.47 \\
        Annif          & 0.66 & 0.28 & 0.56 & 0.46 & 0.5  & Annif          & 0.46 & 0.3  & 0.33 & 0.42 & 0.45 \\
        TartuNLP       & 0.63 & 0.26 & 0.55 & 0.44 & 0.49 & Jim            & 0.4  & 0.29 & 0.29 & 0.39 & 0.43 \\
        Jim            & 0.62 & 0.29 & 0.5  & 0.44 & 0.49 & icip           & 0.39 & 0.28 & 0.3  & 0.4  & 0.42 \\
        icip           & 0.57 & 0.27 & 0.48 & 0.43 & 0.48 & TartuNLP       & 0.4  & 0.26 & 0.31 & 0.38 & 0.41 \\
        LA$^2$I$^2$F   & 0.52 & 0.25 & 0.43 & 0.39 & 0.46 & LA$^2$I$^2$F   & 0.34 & 0.25 & 0.25 & 0.35 & 0.4  \\
        NBF            & 0.44 & 0.21 & 0.41 & 0.37 & 0.43 & NBF            & 0.23 & 0.17 & 0.2  & 0.28 & 0.32 \\
        last\_minute   & 0.27 & 0.15 & 0.24 & 0.25 & 0.29 & Homa           & 0.19 & 0.15 & 0.15 & 0.21 & 0.22 \\
        Homa           & 0.3  & 0.16 & 0.25 & 0.26 & 0.27 & last\_minute   & 0.13 & 0.1  & 0.11 & 0.18 & 0.2  \\
        YNU-HPCC       & 0.21 & 0.14 & 0.18 & 0.22 & 0.26 & YNU-HPCC       & 0.12 & 0.1  & 0.1  & 0.16 & 0.17 \\
        \bottomrule
    \end{tabular}
    \caption{Qualitative performance comparison across teams for two cases: case 1 — treating both Y and I as correct, and case 2 — treating only Y as correct. Metrics reported are precision and recall at $k$. The `Ov. R@k' columns represent the average recall across $k = 5, 10, 15, 20$.}
    \label{tab:qual-team-results}
\end{table*}

\begin{figure*}[!htb]
  \includegraphics[width=\linewidth]{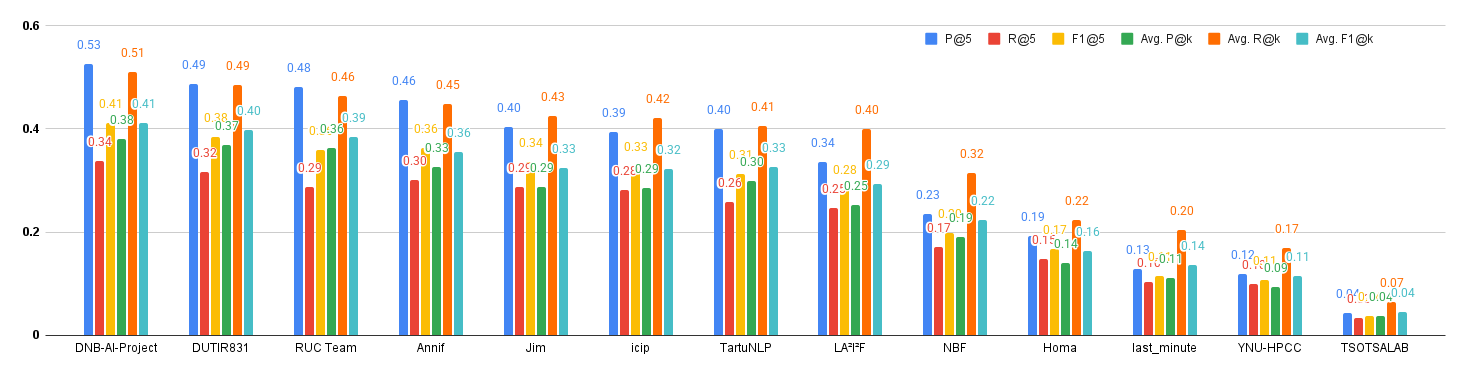}
  \caption{Overall qualitative evaluation results w.r.t. metric@5 and averages per metric@k where k = 5, 10, 15, and 20. On the x-axis, teams are listed in ranked order of performance based on average recall@k.}
  \label{fig:qual-overall-results}
\end{figure*}

\begin{figure*}[!htb]
    \centering
    \small
    \begin{subfigure}{\textwidth}
        \centering
        \includegraphics[width=\linewidth]{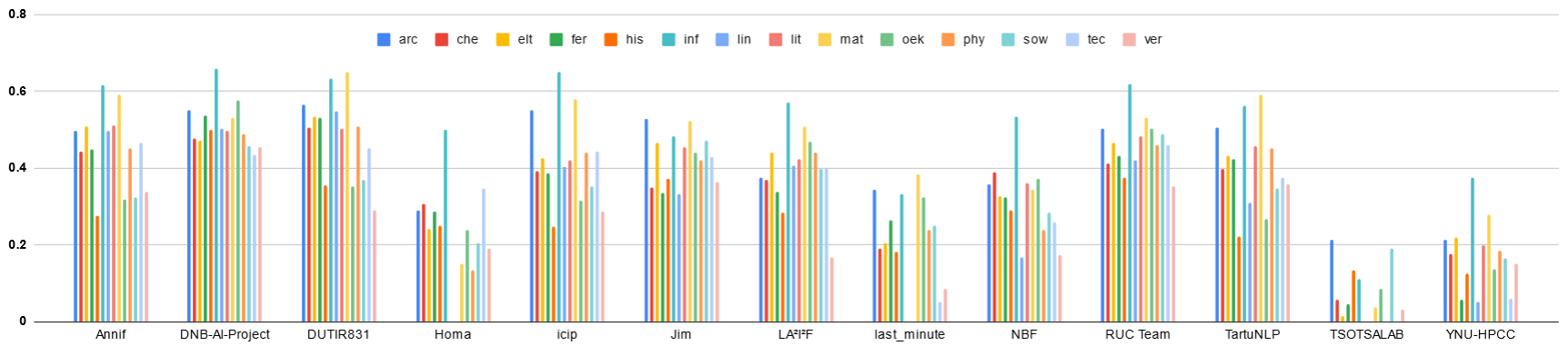}
        \caption{Average recall@k scores per domain over k = 5, 10, 15, and 20.}
        \label{fig:rec-perdomain}
    \end{subfigure}

    \bigskip

    \begin{subfigure}{\textwidth}
        \centering
        \includegraphics[width=\linewidth]{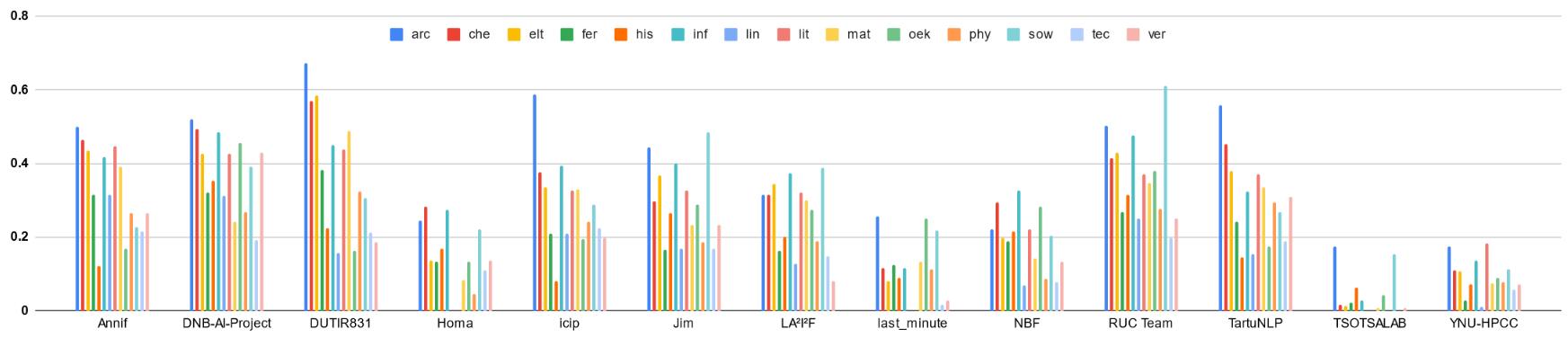}
        \caption{Average precision@k scores per domain over k = 5, 10, 15, and 20.}
        \label{fig:prec-perdomain}
    \end{subfigure}

    \caption{Qualitative results per 14 distinct domains. Acronyms used: Architecture (arc), Chemistry (che), Electrical Engineering (elt), Material Science (fer), History (his), Computer Science (inf), Linguistics (lin), Literature Studies (lit), Mathematics (mat), Economics (oek), Physics (phy), Social Sciences (sow), Engineering (tec), and Traffic Engineering (ver). On the x-axis, teams are listed in alphabetical order of names.}
    \label{fig:qual-perdomain}
\end{figure*}

\subsection{Qualitative Evaluations}

We now turn to the qualitative manual evaluations from the shared task’s evaluation phase. 

\autoref{tab:qual-team-results} shows results for both qualitative evaluation cases. Based on subject librarian assessments, both Y (correct) and I (irrelevant but technically correct) labels were counted as correct in case 1, while only Y was considered correct in case 2. The top 4 teams ranked consistently across both cases, with minor changes among the remaining teams. Case 1 accounts for situations where models predicted multiple semantically similar subjects as top-ranked, leading to generally higher precision scores. However, in practice, it is preferred that models predict semantically distinct and relevant subjects as top-ranked. Therefore, the remainder of this section focuses on case 2, where only Y labels are treated as correct.

The overall results, shown in \autoref{fig:qual-overall-results}, report six metrics: P@5, R@5, F1@5, Avg. P@k, Avg. R@k, and Avg. F1@k, with $k$ ranging from 5 to 20 in the qualitative setting. These results do not distinguish between all-subjects and tib-core since the 122 evaluated records were shared across both collections.\footnote{The silp\_nlp team output was not evaluated qualitatively as it was submitted after the deadline.} The top teams from the quantitative leaderboard (DUTIR831, RUC Team, and Annif) remained among the top four, with the DNB-AI-Project emerging as the best-performing system in qualitative evaluations. Here, precision reflected true system performance, measuring the proportion of predicted subjects marked correct by subject librarians. Recall was adjusted to account for any newly identified correct subjects not present in the gold standard. DNB-AI-Project stood out for its high precision among the top 20 recalled subjects, employing a purely LLM-based approach with an ensemble of LLMs and few-shot prompting, requiring no fine-tuning. This supports the premise of the shared task—assessing whether LLMs can generalize effectively compared to traditional machine learning approaches that rely on extensive fine-tuning. Among the 14 evaluated domains, Computer Science (inf) consistently had the highest average recall (\autoref{fig:rec-perdomain}), while Linguistics (lin) and Literature Studies (lit) showed no predictions from Homa and last\_minute. Most teams struggled with Engineering (tec) and Traffic Engineering (ver) records, and Annif and DUTIR831 also exhibited low recall for History (his) and Economics (oek). In contrast, the DNB-AI-Project and RUC Team demonstrated consistent performance across all domains. Finally, as shown in \autoref{fig:prec-perdomain}, precision did not always align with recall rankings; the Architecture (arc) domain, however, exhibited the top 2 highest precision among all domains.

\section{Discussion}

To establish a reference point for the LLMs4Subjects shared task, we developed a  \href{https://github.com/jd-coderepos/llms4subjects/tree/main/shared-task-baseline}{baseline system} using OpenAI’s GPT-4o via the Assistant API.\footnote{\url{https://platform.openai.com/docs/api-reference/assistants}} Two assistants were configured—one for all-subjects and another for tib-core—each equipped with the respective GND subject taxonomies stored in OpenAI’s \href{https://platform.openai.com/docs/api-reference/vector-stores}{vector stores}. For each TIBKAT record, the assistants embedded the title and abstract and queried the vector store to retrieve 50 GND subjects.

Both assistants followed an identical \href{https://github.com/jd-coderepos/llms4subjects/blob/main/shared-task-baseline/prompts/all-subject-assistant.txt}{prompt} that defined their role as a subject matter expert in a technical library using the GND taxonomy. The prompt instructed them to select exactly 50 valid, semantically relevant subject tags based on the input title and abstract, and return them in a strict JSON format. It also enforced constraints such as avoiding duplicates and non-matching entries, and supported bilingual input in German and English.

Despite using a GPT LLM and a well-structured prompt, the Assistant API presented reliability issues: for some records, the output JSON was inconsistent with the schema, or contained improperly formatted GND codes, requiring repeated runs to obtain usable results. The Assistant API is still in beta, and its stability at scale remains uncertain. This baseline ranked below participant submissions as shown on the \href{https://sites.google.com/view/llms4subjects/team-results-leaderboard}{leaderboard}. These findings highlight that while a general-purpose LLM with vector retrieval and prompt engineering—accessed via the Assistant API interface—offers quick prototyping, effective subject tagging requires more specialized and robust approaches, as shown by the top-performing teams.

\section{Conclusion}

SemEval-2025 Task 5: LLMs4Subjects presented the first evaluation of LLMs for GND-based subject indexing in bilingual (German/English) technical library records, combining quantitative metrics with expert assessments. The task revealed four key takeaways: multilingual models and training data outperformed monolingual ones (RQ1, RQ2); synthetic data and retrieval-augmented pipelines improved performance, underscoring the value of data diversity and system design (RQ3); and smaller, well-engineered systems often rivaled large instruction-tuned LLMs, highlighting trade-offs between scale and specialization (RQ4).

All data, code, and evaluation resources are openly available at \url{https://github.com/jd-coderepos/llms4subjects}. A second edition\footnote{\url{https://sites.google.com/view/llms4subjects-germeval/}} of the task is planned, with an emphasis on \textit{energy- and compute-efficient} LLM systems.

\section*{Acknowledgments}

The SemEval Task-5 LLMs4Subjects shared task, is jointly supported by the TIB Leibniz Information Centre for Science and Technology, the \href{https://scinext-project.github.io/}{SCINEXT project} (BMBF, German Federal Ministry of Education and Research, Grant ID: 01lS22070), and the \href{https://www.nfdi4datascience.de/}{NFDI4DataScience initiative} (DFG, German Research Foundation, Grant ID: 460234259).

\bibliography{custom}

\appendix

\section{TIBKAT media types}
\label{sec:mt-def}

    \noindent{\textbf{Book}}: A comprehensive written work published as a volume or a series of volumes.
    
    \noindent{\textbf{Thesis}}: A document submitted in support of candidature for an academic degree or professional qualification, presenting the author's research and findings.
    
    \noindent{\textbf{Conference}}: Proceedings or collections of papers presented at academic conferences or symposiums.
    
    \noindent{\textbf{Report}}: Detailed and systematic accounts of research findings, often prepared for a specific audience or purpose.
    
    \noindent{\textbf{Article}}: A written composition on a specific topic, usually intended for publication in a journal or magazine.
    Collection: A curated assembly of documents or works, typically related by theme, subject, or author.
    
    \noindent{\textbf{AudioVisualDocument}}: Media content that combines both sound and visual components, such as videos or films.
    
    \noindent{\textbf{Periodical}}: Publications issued at regular intervals, such as journals, magazines, or newspapers.
    
    \noindent{\textbf{Chapter}}: A specific section or segment of a book, usually focusing on a particular topic within the larger work.

\section{TIBKAT Language Distribution}

There are 48 different languages in the TIBKAT with abstracts. Their distributions are listed below.

de (German): 108637; en (English): 76735; fr (French): 1741; id (Indonesian): 945; es (Spanish): 311; it (Italian): 294; nl (Dutch): 167; da (Danish): 129; sq (Albanian): 100; ro (Romanian): 93; ca (Catalan): 86; fi (Finnish): 80; so (Somali): 67; sv (Swedish): 65; no (Norwegian): 50; unknown (Unknown): 41; tl (Tagalog): 31; et (Estonian): 24; pt (Portuguese): 16; sw (Swahili (macrolanguage)): 15; pl (Polish): 15; hr (Croatian): 12; lt (Lithuanian): 10; hu (Hungarian): 10; af (Afrikaans): 10; tr (Turkish): 8; sk (Slovak): 7; sl (Slovenian): 4; cy (Welsh): 4; cs (Czech): 3; lv (Latvian): 3; vi (Vietnamese): 2; he (Hebrew): 2; ko (Korean): 1; ru (Russian): 1.




\section{Additional details about the GND}

At TIB, the GND is used as follows. The subject specialists are usually using online services like GND-Explorer (\url{https://explore.gnd.network/}) or OGND (\url{https://swb.bsz-bw.de/}) for searching the GND. There you can also restrict to Sachbegriff/Topical term (saz), which is the term class we are using for subject indexing (there are some additional term classes used, like geographical terms (swg), but to start topical terms (saz) should be sufficient). New terms are added in a cooperative process and are searchable as soon as they passed a review process. For more general details you can also have a closer look \href{https://www.dnb.de/EN/Professionell/Standardisierung/GND/gnd_node.html}{here}. There is also a way to get complete sets of the GND that are updated on a regular basis. Details can be found \href{https://www.dnb.de/EN/Professionell/Metadatendienste/Datenbezug/Gesamtabzuege/gesamtabzuege_node.html}{here}. A centralized resource pool and a guide for accessing the GND are provided in the LLMs4Subjects GitHub repository \url{https://github.com/jd-coderepos/llms4subjects/tree/main/gnd-how-to}.

Note, all terms in GND usually are in German and every TIB record regardless of its language is described by it. But there are some cases, where a term is e.g. English, if the preferred naming is English. In this case a German naming can be listed under synonyms. The synonyms are also important for the subject classification, as many relevant terms are listed under synonyms. These are not always synonyms in a strong sense, as the GND is a growing database and meanings of terms once created do change or shift and larger corrections are rarely realized, if terms are not wrong in sense of content. The GND's purpose extends to enhancing the discoverability of literature in bibliographic systems, where synonyms are also utilized, emphasizing their importance in indexing.

\section{Pilot Task}
\label{pilot-task}

At TIB, ANNIF has been used in production code since the start of 2024 for the use case of assigning items from the TIB portal discovery system to one or several subject facets.
The TIB portal employs a multi-stage algorithm to attribute a record to one of the 28 TIB's different subjects, viz. Architecture, Civil Engineering, Biochemistry, Biology, Chemistry, Chemical Engineering, Electrical Engineering, Energy Technology, Educational Science, Earth Sciences, History, Information Technology, Literary Studies and Linguistics, Mechanical Engineering, Mathematics, Medical Technology, Plant Sciences, Philosophy, Physics, Law, Study of Religions, Social Sciences, Sports Sciences, Theology, Environmental Engineering, Traffic Engineering, Materials Science, and Economics, the last of which is the so-called automatic stage.
If no more salient information is available, machine learning methods are used to assign the subject(s).
Note, the subjects reference here can be seen directly akin to fields of study or scientific disciplines, whereas LLMs4Subjects includes a much broader scope for its subjects.
Previously utilizing a commercial algorithm, TIB switched to ANNIF for its customization potential and community-driven improvements.
The training data of ANNIF algorithms consists of document metadata from the TIB catalog, partially overlapping with the training dataset for LLMs4Subjects.
Since the documents to be indexed by ANNIF include many cases where abstracts or fulltexts cannot be accessed programmatically, we only consider the the titles and publishers.
ANNIF has shown good overall results in assigning the 14 subjects is has so far been incrementally trained on, with an overall F1 score of $\approx\!0.65$ for several algorithms.
Both English and German-language documents were considered, with little difference in performance when training both languages combined or separately.
Leveraging the capabilities of LLMs as a complementary approach to ANNIF marks a logical next step in the automation of subject indexing.

\end{document}